\documentclass[11pt,a4paper]{article}
\usepackage[hyperref]{emnlp-ijcnlp-2019}
\usepackage{times}
\usepackage{latexsym}

\usepackage{helvet}
\usepackage{courier}
\usepackage{graphicx}
\usepackage{amsmath}
\usepackage{latexsym}
\usepackage{amssymb}
\usepackage{multirow}
\usepackage{amssymb}
\usepackage{amsmath}
\usepackage{setspace}
\usepackage{booktabs}
\usepackage{CJK}
\usepackage{enumitem}
\usepackage{array}
\usepackage{url}

\aclfinalcopy % Uncomment this line for the final submission

\title{Unsupervised Word Segmentation with  Bi-directional \\ Neural Language Model}

\author{
Lihao Wang, Zongyi Li, Xiaoqing Zheng \\
School of Computer Science, Fudan University, Shanghai, China \\
Shanghai Key Laboratory of Intelligent Information Processing \\
\texttt{\{wanglh19, zongyili19, zhengxq\}@fudan.edu.cn }
}

\date{}

\begin{document}
\begin{CJK}{UTF8}{gbsn}

\maketitle
\begin{abstract}
We present an unsupervised word segmentation model, in which the learning objective is to maximize the generation probability of a sentence given its all possible segmentation. Such generation probability can be factorized into the likelihood of each possible segment given the context in a recursive way. In order to better capture the long- and short-term dependencies, we propose to use bi-directional neural language models to better capture the features of segment's context. Two decoding algorithms are also described to combine the context features from both directions to generate the final segmentation, which helps to reconcile word boundary ambiguities. Experimental results showed that our context-sensitive unsupervised segmentation model achieved state-of-the-art at different evaluation settings on various data sets for Chinese, and the comparable result for Thai. 
\end{abstract}

\section{Introduction}
Word segmentation aims to split a natural language sentence into its component words. 
In English and many other Latin languages, whitespace is used as a natural delimiter among words, while in character-based language such as Chinese and Japanese, we usually need to recover implicit word boundaries before further processing on language tasks.
Previous psychological studies \cite{cattell1886time, li2009segmentation} indicate that words are processed in a word-level way instead of character-level way, which means word segmentation is an essential step for further researches.
% \begin{figure}
% 	\centering
% 	\includegraphics[width=0.5\textwidth]{sample_2.pdf}
% 	\caption{Example sentences where the context features from both directions are required to reconcile the word boundary ambiguities.}
% 	\label{Figure9}
% \end{figure}

\begin{table}[!t]
\begin{center}
	\caption{Example sentences where the context features from both directions are required to reconcile the word boundary ambiguities.}
    \begin{tabular}{l|c}
    \hline
    \hline
     我 / 从小 / 学 / 唱歌 & \multicolumn{1}{m{3.6cm}}{I started learning to sing when I was young.} \\ 
     \hline
     我 / 从 / 小学 / 毕业 & \multicolumn{1}{m{3.6cm}}{I graduated from primary school.} \\ 
    \hline
    \hline
    \end{tabular}
    \label{sample}
\end{center}
	\end{table}
There are two main learning paradigms for word segmentation: namely supervised and unsupervised approaches.
Although supervised methods have generally achieved higher accuracy \cite{huang2007chinese}, these models need a large volume of manually labeled corpora to be trained, which requires expensive human efforts.
Furthermore, supervised approaches often train domain-specific models, which may achieve accurate prediction on texts in the same domain as the training texts, but significantly deteriorate performance when applied to texts in other domains.
Moreover, there is still no segmentation standard widely accepted for some languages.
For example, in Chinese, the meaning of a multi-word phrase may vary greatly if the words are considered separately or together.
Unsupervised word segmentation approaches can deal with the three issues mentioned above.
The training with unlabeled dataset requires much less human effort, and previous studies show that unsupervised learning has better stability and adaptivity to novel domain, such as raw texts with unseen words or in new languages.
All these advantages over supervised learning makes unsupervised approaches worth further investigation. 

Traditional work on unsupervised word segmentation can be divided into two categories: rule-based approaches and statistical approaches. 
The rule-based approaches identify words by applying prior knowledge or morpho-lexical rules governing the derivation of  \cite{sproat1990statistical,jin2006unsupervised,magistry2012unsupervized}, while the statistical approaches identify words based on the distribution of their components in a larger corpus \cite{teh2006hierarchical,goldwater2009bayesian,mochihashi2009bayesian,uchiumi2015inducing}.
The former is intrinsically context-free, so it could not deal with the word boundary ambiguities (i.e. combination and overlap ambiguity). The hyper-parameters of the latter need to be carefully tuned for a training corpus, which might not be still work for others.

Recently, neural networks have been tried to address the problem of unsupervised word segmentation.
\citet{wang2017sequence} proposed the Sleep-WAke Network (SWAN) for speech recognition and also apply it to text segmentation.
\citet{sun2018unsupervised} extended SWAN model for Chinese word segmentation and presented a Segmental language models (SLM).
However, those models only consider the context from a direction (left-to-right) and can not well resolve the word boundary ambiguities that only can be reconciled by using the context features from both directions (see example Chinese sentences listed in Table \ref{sample}).

% etc.

We propose an unsupervised word segmentation model based on a bi-directional neural language model by maximizing the generation probability of sentences, which can be recursively factorized into the likelihood of their all possible segmentation, taking their rich-context features into consideration.
Two decoding algorithms were proposed to combine the context features from both directions. One is to produce the segmentation by taking the average of segment's probabilities estimated by the forward and backward language models. Another is to generate the segmentation by merging their results produced separately by those two models. The former revealed better in word boundary ambiguity reconciliation, while the latter achieved the higher overall performance.  
% We come up with the following two methods to take full advantage of our bi-directional (forward and backward) neural model:
% averaging the two probabilities got from our model to represent a certain context-sensitive segment probability, and then using dynamic decoding algorithm to decode the best segmentation;
% decoding forward and backward best segmentation independently and then integrating the result by keeping all the cuts.
% By experiments, the second approach achieves higher accuracy, while the first one has better disambiguation ability.

Our model can successfully split a sentence into words without any prior knowledge or manually designed features.
The bi-directional language models are used to better capture the long- and short-term dependencies within sentences.
% leading to a noticeable improvement in word segmentation.
Experimental results on Chinese datasets showed that our model performed better than SWAN \cite{wang2017sequence} and SLM \cite{sun2018unsupervised} with a significant margin.
Our context-sensitive unsupervised model achieved state-of-the-art on multiple Chinese benchmarks with different settings.
We also evaluated our model on Thai in which the words are formed by more characters in average than those in Chinese, and achieved comparable result, comparing with recent strong competitors.

% The rest of the paper is organized as follows.
% Section 2 presents a brief overview of related work.
% Section 3 presents our unsupervised framework and the neural network architecture implemented in the framework. Section 4 reports the experimental results of our model by comparing to some representative systems and future work are given in section 5.

\section{Related Work}

\subsection{Traditional Methods}
There are two main paradigms for the traditional works of unsupervised Chinese word segmentation: one uses goodness measures to find boundaries between words, while the other estimates the probability by nonparametric Bayesian models.

\citet{sproat1990statistical} first applied mutual information (MI) to word segmentation, and then many methods have been proposed based on the variants of goodness measures, including cohesion measure, separation measure and a combination of two.
Several goodness measures were compared by \citet{zhao2008empirical}.
They propose a unified framework, including description length gain (DLG) \cite{kitt1999unsupervised}, accessor variety (AV) \cite{feng2004accessor}, and branch entropy (BE) \cite{jin2006unsupervised}.
\citet{magistry2012unsupervized} promoted on the branch entropy method, and proposed nVBE by adding normalization and viterbi decoding which removed most hyper-parameters.
% Since then, many works are based on the combination of the above.
% The Minimum Description Length (MDL) introduced by \citet{rissanen1978modeling} is often used in unsupervised segmentation systems to select optimal parameter, and search for a more compact solution from candidate segmentations.
% \citet{zhao2008empirical} used MDL to process two-character words and AV to longer. 
% \citet{hewlett2011fully} present Bootstrap Voting Expert (BVE) which combines the votes of different indicators based on BE, and used MDL to improve the segmentation.
\citet{wang2011new} proposed ESA (Evaluation, Selection, Adjustment), which used goodness measures to adopt a local maximum strategy without thresholds.

The other important paradigm used in unsupervised word segmentation is nonparametric Bayesian model.
Dirichlet process (DP) \cite{maceachern1998estimating}, Pitman-Yor process (PYP) \cite{pitman1997two}, hierarchical DP (HDP)  \cite{teh2005sharing, goldwater2006contextual}, hierarchical PYP (HPYP) \cite{teh2006hierarchical} have been introduced to word segmentation.
\citet{goldwater2009bayesian} proposed a unigram and bigram model for unsupervised word segmentation, with the help of DP and HDP.
However, these approaches are computationally intensive.
\citet{mochihashi2009bayesian} introduced a faster algorithm based on a nested Pitman-Yor language model (NPYLM) by introducing a nested character model and an efficient blocked Gibbs sampler combined with dynamic programming for inference and Pitman-Yor Hidden Semi-Markov Model (PYHSMM) \shortcite{uchiumi2015inducing} which was considered as a method to build a class n-gram
language model directly from strings while integrating character and word level information.
Their evaluation results outperformed the previous.
Joint model of hierarchical Dirichlet process model and character-based hidden Markov model \citep{chen2014joint} and Bayesian model based on monolingual character \citep{teng2014unsupervised} achieved state-of-art performance in different datasets.

% Goodness measure-based models can hardly disambiguate words because they are context-free, thus segmenting ambiguous strings into the same word sequences.
% Non-parametric Bayesian models tending to segment the sequences into sentences that are more reasonable, alleviate the above problems to some extent, while introducing new problems of adaptability and computation ability, because they need to adjust a large number of hyper-parameters and prior distribution.

\subsection{Neural Methods}
% Neural networks have been applied to word segmentation these years whose performance have overtaken statistical models.
% Recursive neural network (RNN) and convolutional neural network (CNN) have been applied to supervised Chinese word segmentation, whose performance has overtaken statistical models.

% Semi-supervised training strategies have been used to take advantage of unlabeled data to enhance the performance.
% For example, \citet{peters2017semi} introduced bidirectional language model to improve sequence labeling tasks.
% Recently, some segmental sequence model have been proposed to segmental labeling.
\citet{kong2015segmental} introduced segmental recurrent neural networks (SRNNs) which define a joint probability distribution over segmentations of the input and labeling of the segments.
This model can be partially training in which segment boundaries are latent, but can not extend the task fully unsupervised in the case of neither labels nor latent variable.
\citet{wang2017sequence} proposed the Sleep-WAke Network (SWAN) which sum over all valid segmentations to obtain the final probability for the sequence.
SWAN model present a sentence-level loss function under unsupervised conditions.
Aimed for speech recognition, SWAN is a online segment to segment model which used beam search decoding algorithm.
\citet{sun2018unsupervised} extended SWAN to Chinese word segmentation and use dynamic decoding algorithm to get best segmentation path.
Neither Wang nor Sun consider the influence of backward context information on segmentation.  

\section{Methodology}
We assume that the result of word segmentation can be determined by a latent variable, and its prior distribution is a uniform distribution over all the possible segmentations for a sentence.
We require that a good segmentation should contribute to increase the generation probability of a sentence, and thus we can choose the segmentation with the highest generation probability for the sentence as the final result.
Our unsupervised segmentation model learns to maximize the generation probability of each sentence given its all the possible word segmentations.
% can be estimated by maximum likelihood and we find the learning objective of the likelihood function is to maximize the generation probability of sentences given its all possible segmentation.
We show that such probability can be factorize into the likelihood of each possible segment (or fragment) of a sentence given its context in a recursive way.
Bi-directional neural language models are used to capture the context information for a segment and to estimate the probabilities of the segment from the both sides.
% We combine two-side probabilities by averaging them to make full use of the context information. 
Two dynamic decoding algorithms are proposed to combine the context information from the both sizes to produce the best segmentation.

\subsection{Unsupervised Framework}
We introduce a latent variable $z$ that determines the result of segmentation. If we have a sentence denoted as $d$, then the probability of $z$ given $d$ with parameters $\theta$ can be formalized as follows:
\begin{equation}\small
p_{\theta}(z|d)=\frac{p_{\theta}(d|z)p_{\theta}(z)}{p_{\theta}(d)}
\end{equation}
Since $p_{\theta}(d)$ is a constant term, we can find the best segmentation by finding the value of $z$ that maximizes $p_{\theta}(d|z)p_{\theta}(z)$. 
% $p_{\theta}(d|z)$ denotes the likelihood of observed data given the certain segmentation which can be generated from the segmentation models.
It is usually easier to estimate $p_{\theta}(d|z)$ by proposing a model that can be used to generate the observed data $d$.
If we assume the prior of $p_{\theta}(z)$ is a uniform distribution, 
then we can gain the best segmentation $z^{*}$ via:
\begin{equation}\small z^{*}=\mathop{\arg\max}_{z} p_{\theta}(d|z)\end{equation}
We optimize the model's parameters $\theta$ by maximizing a log-likelihood function $L$. If marginalizing all the segmentation $z$, we have:
\begin{equation}\small 
L(\theta)=\sum\limits_{d}\ln p_{\theta}(d)=\sum\limits_{d}\ln\sum\limits_{z}p_{\theta}(d|z)
\end{equation}
In the above equation, we need to compute the sum of generation probabilities for a sentence over all the possible segmentations. However, it is infeasible to enumerate all the possible segmentations, and so we use dynamic programming technique to compute such probability.

\subsection{Generation Probability}
A forward-backward algorithm is an inference algorithm used in hidden Markov models, which computes the posterior marginals of all hidden state variables given a sequence of observations\cite{rabiner1989tutorial}. In our method, the algorithm can be applied to calculate the sum probability of the sentences generated by all possible segmentation. 
% According to Markov property, the calculated probabilities are related to the previous states(previous segmentations and preceding context)

We follow the symbol of \citet{mochihashi2009bayesian}.
First, we define $\alpha$ as forward variable, $t$ as the character sequence length, $k$ as the predicted word length and abbreviated character sequence $c_{m}\cdots c_{n}$ to $c_{m:n}$.
Then, the forward variable $\alpha_{t}^{k}$ means the probability of character sequence $c_{1:t}$ with final $k$ characters being a word. 
We also use $q_{t}=k$ to represent the occurrence of a sequence of length $t$ whose final $k$ characters is a word. Then, we have the following deduction: 

\begin{equation}\small
 \begin{split}
 \alpha_{t}^{k} & = p(c_{1:t},q_{t}=k) \\
 & = p(c_{1:t-k},c_{t-k+1:t},q_{t}=k) \\
 & = \sum\limits_{j=1}^{t-k}p(q_{t}=k,c_{t-k+1:t},c_{1:t-k},q_{t-k}=j)\\
 & = \sum\limits_{j=1}^{t-k}p(q_{t}=k|c_{t-k+1:t})p(c_{t-k+1:t}|c_{1:t-k})\alpha_{t-k}^{j}
  \end{split}
\end{equation}
 
Note that $p(c_{t-k+1:t} | c_{1:t-k})$ denotes the generated probability of segment character sequence given the preceding context and $p(q_{t}=k | c_{t-k-1:t})$ denotes the probability of the segment being a word. These two kinds of probability will be modeled in next section. 
Now, we can calculate the sum probability via forward variable :
\begin{equation}\small 
\sum\limits_{z}p_{\theta}(d | z)=\sum\limits_{k=0}^N\alpha_{N}^{k}
\end{equation}
where $N$ is the length of the sentence and $\alpha_{0}^{0}=1$

According to Markov property, the probabilities above are related to the previous states(previous segmentations and preceding context). In order to leverage the following context to segmentation, we  define a backward variable as $\beta_{t}^{k}$ and just reverse the sentences and apply the equation (4) again to obtain another sum probability. 
The  probabilities calculated by forward variable $\alpha$ denote the  generative probabilities with preceding context , while the probability calculated by backward variable $\beta$ denotes the generative probability with following.
We average the above two probabilities  to make full use of the context information of both sides.
Now we can obtain the objective function $J(\theta)$ according to equation (4)-(6):

\begin{equation}\small
 \begin{split}
 J(\theta) & = -L(\theta) \\
& = - \sum\limits_{d}\ln\sum\limits_{z}p_{\theta}(d|z)\\
& = - \frac{1}{2}\sum\limits_{d}\ln(\sum\limits_{k=0}^{N_{d}}\alpha_{N_{d}}^{k}\sum\limits_{k=0}^{N_{d}}\beta_{N_{d}}^{k})
  \end{split}
\end{equation}

\subsection{Model}
In the previous two sections, we present an unsupervised framework to inference the best segmentation $z$, and apply max-likelihood to estimate model parameter $\theta$ as well.
We also factorized the objective function into the probability modeling of $p(q_{t}=k|c_{t-k+1:t})$ and $p(c_{t-k+1:t}|c_{1:t-k})$. In this section, we will utilize neural language model to generate these probabilities.
\subsubsection{Neural Network-based Implementation}
Now, we introduce a neural network architecture to implement the above framework.

Firstly, $p(c_{t-k+1:t}|c_{1:t-k})$ can be further factorized and generated by neural language models. We also denote the possible segment character sequence $c_{t-k+1:t}$ as $s_{1:k}$  :
\begin{equation}\small
p(c_{t-k+1:t}|c_{1:t-k})=\prod_{i=1}^{k}p(s_{i}|s_{1:i-1},c_{1:t-k})
\end{equation}
Recent work \cite{bengio2003neural, mikolov2010recurrent, jozefowicz2016exploring} suggests that recurrent neural network (RNN), and its variants Long short-term memory (LSTM) \cite{hochreiter1997long}, Gated Recurrent Unit (GRU) \cite{cho2014learning} have a good ability to model language models, as they have the ability to capture the long- and short-term dependencies.
Therefore, we use four LSTMs to obtain the probabilities : two for generating context representations, named as context LSTMs, two as neural language models to generate above probabilities, named as language model LSTMs , each aspect can be split into two directions, forward and backward.

We take the forward direction as an example, the forward context LSTM is used to capture the preceding context information.Given a N-length character embedding input sequence of sentences $(c_{1}\cdots c_{N})$, the context LSTM $f^{C }$ produce the forward hidden sequence $(\stackrel{\rightarrow}{\mathbf{h_{1}^{C}}} \cdots \stackrel{\rightarrow}{\mathbf{h_{N}^{C}}})$ as representations of the preceding context, which will also be used as initial hidden state of a forward language model LSTM. At each time step $t$, the hidden state $\mathbf{h_{t}^{C}}$ of the context LSTM is updated
by:
\begin{equation}\small
\mathbf{h}_{\mathbf{t}}^{\mathbf{C}}=f^{C}\left(\mathbf{h}_{\mathbf{t}-\mathbf{1}}^{\mathbf{C}}, c_{t-1}\right)
\end{equation}
Then, the language model LSTM $f^{L}$ can learn a probability distribution
over a possible segment character  sequence $s_{1:k}$  by being trained to predict the
next character in the sequence via softmax function. The initial hidden state $\mathbf{h_{0}^{L}}$ of the language model LSTM obtains from hidden sequence generated by the context LSTM at the time step $t-k+1$ , which represents the preceding context of sequence $s_{1:k}$.  :
\begin{equation}\small
\mathbf{h}_{0}^{\mathbf{L}}=\mathbf{h}_{\mathbf{t}-\mathbf{k}+1}^{\mathbf{C}}
\end{equation}

\begin{equation}\small
\mathbf{h}_{\mathbf{t}^{\prime}}^{\mathbf{L}}=f^{L}\left(\mathbf{h}_{\mathbf{t}^{\prime}-\mathbf{1}}^{\mathbf{L}}, s_{t^{\prime}-1}\right)
\end{equation}

\begin{equation}\small
p(s_{i}|s_{1:i-1},c_{1:t-k})=\frac{\exp \left(\mathbf{w}_{j} \mathbf{h}_{i}^{L}\right)}{\sum_{j^{\prime}=1}^{V} \exp \left(\mathbf{w}_{j^{\prime}} \mathbf{h}_{i}^{L}\right)}
\end{equation}

Finally, we combine these probability using equation (7) to obtain the segment probability with preceding context.
Similarly, the backward context LSTM computes the backward hidden sequence $(\stackrel{\leftarrow}{\mathbf{h_{1}^{C}}} \cdots \stackrel{\leftarrow}{\mathbf{h_{N}^{C}}})$ as the following context information and the backward language model LSTM generates the probability.

\begin{figure} 
	\centering
	\includegraphics[width=0.5\textwidth]{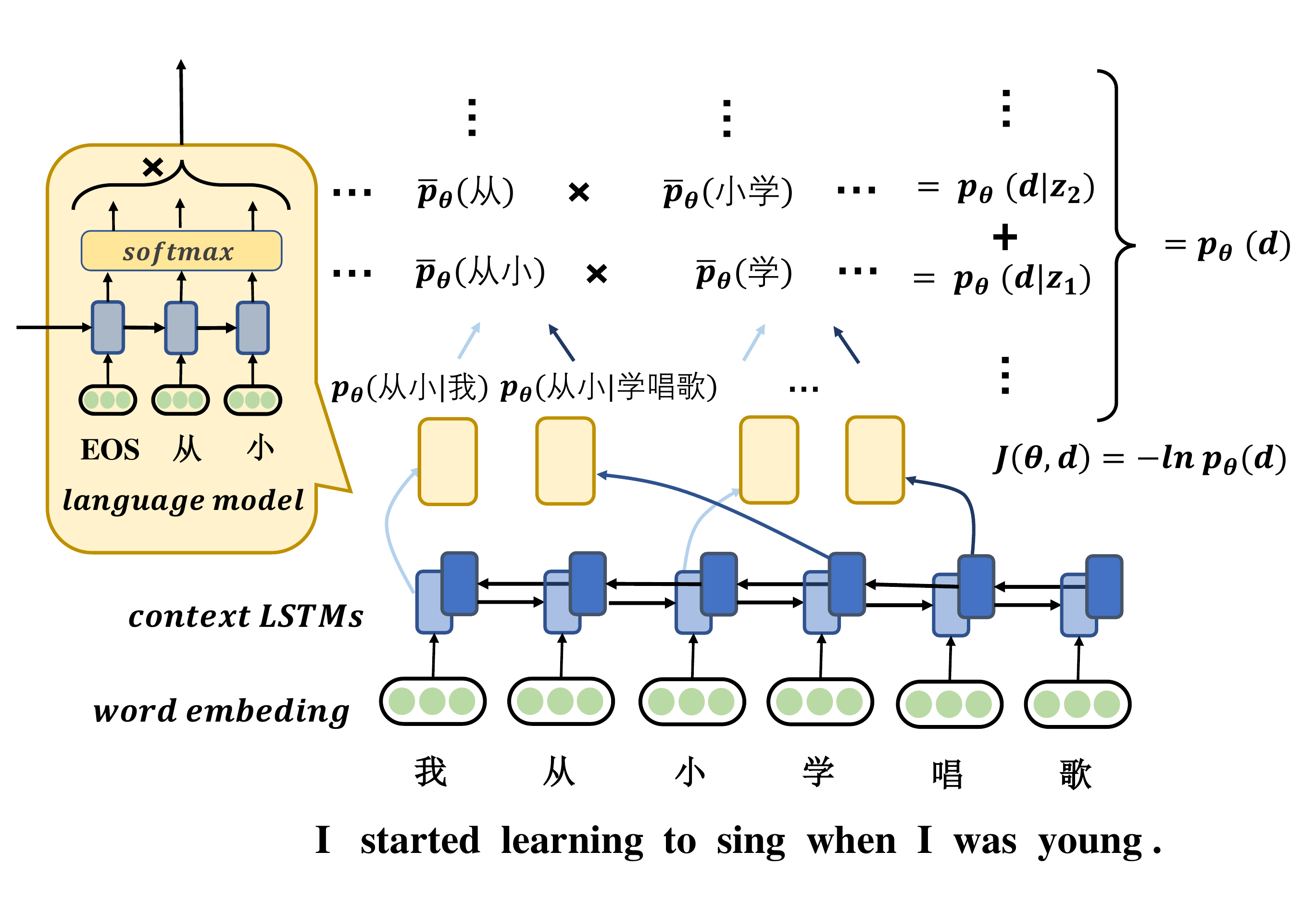}
	\caption{The neural network architecture. In this figure, we want to obtain the probability of possible segment ``从小". Firstly, pick out preceding context representation ``我" and preceding context representation ``学唱歌" from two context LSTMs. Then we used these representations as the initial hidden states of two Language model LSTMs and generate the character probability from both side (We also input the ``EOS" symbol to generate the first character probability). Next , we multiply all the character probabilities to obtain two version of possible segment probabilities. At Last, we average them to get a more accurate probability or utilize them respectively. }
	\label{fig:architecture}
\end{figure}

$p(q_{t}=k|c_{t-k+1:t})$ can be modeled as the probability of an end-of-segment symbol $EOS$,
and the probability $p(q_{t}=k|c_{t-k+1:t})$ is equivalent to $p(EOS|c_{t-k+1:t},c_{1:t-k})$. $EOS$ will compete with all other meaningful characters when optimizing the candidate segment probabilities and reduce the probability of wrong candidate sub-sequences becoming a word.
The architecture of our model is shown in Figure \ref{fig:architecture}.

\subsubsection{Computational Complex and Optimization}
Every candidate segment starting with the same position shares the same forward context, and there is no need to compute it $N$ times.
We store the hidden states of context LSTMs to reduce the algorithm complexity from $O(N^{4})$ to $O(N^{3})$, where $N$ stands for input sequence length.

\citet{yee1986computational} claimed that $69\%$ of Chinese articles consist of single character word, while only $1\%$ consist of three or more character word.
So it is unnecessary to sum all the $k$ (last k character being one word), value from 1 to N.
We set the maximum length of the words to $T(T\geq3)$, reducing the algorithm complexity to $O(N T^{2})$.

In addition, we can compute all the nested sub-sequence probabilities once the generation probability of the entire sequence is obtained. Such probabilities is produced by the softmax layers of our segments.
Therefore, we just need to compute the probability of longest segments, and meanwhile we can get the nested candidate segments' probabilities which share the start position with the longest segments.
We thus can reduce the algorithm complexity from $O(N T^{2})$ to $O(NT)$. 

\subsection{Decoding Algorithms}
In order to simplify the formulation, we denote the product of $p(q_{t}=k|c_{t-k+1:t})$ and $p(c_{t-k+1:t}|c_{1:t-k})$ as the probability $p_{\theta}(w_{i}|ctx(w_{i}))$ of the word $w_{i}$ made up by character sequence $c_{t-k+1:t}$ given its context $ctx(w_{i})$.
The backward version one is symmetrically denoted as $p_{\theta}(w_{i}^{*}|ctx(w_{i}^{*}))$. 
Then, we can compute the probability of a particular segmentation which segments the character sequence $c_{1}\cdots c_{N}$ into the word sequence $w_{1}\cdots w_{M}$ and the reversed character sequence $c_{N}\cdots c_{1}$ into $w_{M}^{*}\cdots w_{1}^{*}$ by multiplying all the generated word probabilities.

We denote $Z(d)$ as the set of all the possible segmentations for the sentence $d$, then we are looking for:
\begin{equation}\small \mathop{\arg\max}_{W,W^{*}\in Z(d)}\prod_{w_{i}\in W}p_{\theta}(w_{i}|ctx(w_{i}))\prod_{w_{i}^{*}\in W^{*}}p_{\theta}(w_{i}^{*}|ctx(w_{i}^{*})\end{equation}
However, it is a NP problem which can not be computed using dynamic programming.
We proposed two dynamic programming methods to approach the solution as follows:\\

\noindent
\textbf{SGB-A } We can combine two probabilities (forward and backward) of a same word by averaging them which is similar to the Figure \ref{fig:architecture} then use dynamic programming algorithm to decode the best segmentation.
We denote $ \hat{\alpha}_{t}^{k}$ as the best segmentation probability of character sequence $c_{1:t}$ with final $k$ characters being a word and $\delta_{t}^{k}$ is used to trace back the decoding.
\begin{equation}\small
    \bar{p}(w_{j})=\sqrt{p_{\theta}(w_{j} | ctx(w_{j}))p_{\theta}(w_{j}^{*} | ctx(w_{j}^{*}))}
\end{equation}
\begin{equation}\small
\hat{\alpha}_{t}^{k}\ =\max\limits_{j=1}^{T}\ \bar{p}(w_{j})\ \hat{\alpha}_{t-k}^{j}
\end{equation}
\begin{equation}\small
\delta_{t}^{k} =\arg\max_{j=1}^{T} \bar{p}(w_{j}) \delta_{t-k}^{j}
\end{equation}
\\
\noindent
\textbf{SGB-C }
We can obtain the best segmentation separately using the two-side probabilities by dynamic programming respectively and integrate all the word boundaries to produce the final segmentation, resulting in a more fine-grained result.

\section{Experiments}
\subsection{Experimental Setting}
\textbf{Datasets } For Chinese word segmentation, we use the second SIGHAN Bakeoff dataset to validate our model.
The second SIGHAN Bakeoff \cite{emerson2005second} have four set of labeled data: AS, CITYU, MSR, PKU. AS and CITYU are written in traditional Chinese, while MSR and PKU are written in simplified Chinese.In similar to most of previous work, we combine the original training data with test data as new training data, and only use test data for evaluation.For Thai word segmentation, InterBest is a dataset in Thai used in the InterBest 2009 word segmentation contest \cite{kosawat2009interbest}.We use the ``Novel" subset for training and test data. The detail is shown in Table \ref{dataset}.\\

\begin{table}[!t]
	\begin{center}
	\caption{Statistics of five corpora.}
		\label{dataset}
		\begin{tabular}{c|c|c|c|c}
			\hline
			\hline
			\multirow{2}{*}{Corpus} & \multicolumn{2}{c|}{Word} & \multicolumn{2}{c}{Character} \\
			\cline{2-5} & Types & Tokens & Types & Tokens  \\
			\hline
			CITYU & 9001 & 40937 & 2953 & 66346 \\
			PKU & 13149 & 104373 & 3433 & 168975 \\
% 			\hline
			MSR & 12923 & 106873 & 3341 & 180987 \\
			AS & 18811 & 122613 & 3884 & 196299 \\
			Novel & 22037 & 1522233 & 156 & 5426551 \\
			\hline
			\hline
		\end{tabular}
	\end{center}

\end{table}
\noindent
\textbf{Preprocess } It is unfair to compare two unsupervised methods under different preprocessing settings.
\citet{wang2011new} proposed four kind of preprocessing settings and we use setting 1, 3 and 4 to compare our model with previous works: 
\begin{itemize} \setlength{\itemsep}{0pt}
\item Setting 1: Fully unsupervised condition.
\item Setting 3: All the punctuation marks are identified as delimiters in advance.
\item Setting 4: All the punctuation marks and other non-Chinese characters are recognized and transformed into special symbols.
\end{itemize}
\noindent
\textbf{Set-up} For our model, we choose LSTM as the language model component and set the embedding size and the hidden size both 300.
Adam algorithm is used for the update rule with learning rate 0.001, and the number of iterates is to 10.
We set different batch sizes in consideration of size of each dataset.
The batch size of CITYU is 32, MSR is 64 , PKU is 16 , AS is 256 and Novel is 256. The vocabulary sizes are equal to size of character set in each datasets.
We test the model with the maximum word length $T = 3, 4, 5$ for Chinese word segmentation and  $T = 12$ for Thai.
We use F1-score on words for evaluation which is the harmonic mean of precision and recall.

\subsection{Result}
We compare our model with some outstanding previous works on five datasets in three settings. The detail information of all models is following: 
% nVBE : a variant of Boundary Entropy method proposed by \citet{magistry2012unsupervized}; ESA : the model proposed by \citet{wang2011new} with best chosen parameters; NPYLM$^{T}$ : $T$-gram Nested Pitman-Yor language model \citep{mochihashi2009bayesian}; PYHSMM : Pitman-Yor Hidden Semi-Markov Model \citep{uchiumi2015inducing}; Joint : Joint model of hierarchical Dirichlet process model and character-based hidden Markov model \citep{chen2014joint}; MCA : Bayesian model based on monolingual character \citep{teng2014unsupervised}; SLM$^{T}$ : segmental language model proposed by \citet{wang2017sequence} and \citet{sun2018unsupervised} with maximum word length $T$.
\begin{itemize}\setlength{\itemsep}{0pt}
\item nVBE: a variant method based on the branch entropy \citet{magistry2012unsupervized}.
\item ESA: the model proposed by \citet{wang2011new} which choose the best value.
\item NPYLM$^{T}$ : $T$-gram Nested Pitman-Yor language model \citep{mochihashi2009bayesian}.
\item PYHSMM: Pitman-Yor Hidden Semi-Markov Model \citep{uchiumi2015inducing}.
\item Joint: Joint model of hierarchical Dirichlet process model and character-based hidden Markov model \citep{chen2014joint}.
\item MCA: Bayesian model based on monolingual character \citep{teng2014unsupervised}.
\item SLM$^{T}$: segmental language model proposed by \citet{wang2017sequence} and \citet{sun2018unsupervised} with maximum word length $T$.
\end{itemize}

\begin{table}[!t]

	\begin{center}

		\caption{F1-score with setting 1 on Chinese word segmentation task.}
			\label{table1}
		\begin{tabular}{l|c|c|c|c}
			\hline
			\hline
			Models (\%)  & CITYU & MSR & PKU & AS\\
			\hline
			ESA & 72.3 & 76.4 & 75.0 & 76.4 \\
			SGB-A$^{3}$  & 78.7 & 79.4 & 78.4 & 79.4 \\
			SGB-C$^{3}$  & 77.4 & 80.2 & 79.6 & 78.6\\
		    SGB-A$^{4}$ & 79.2 & \textbf{80.5} & 77.9 & 80.2 \\
			SGB-C$^{4}$ & \textbf{80.0} & 74.0 & \textbf{80.0} & 81.0 \\
			SGB-A$^{5}$ & 72.5 & 72.8 & 75.4 & 64.5 \\
			SGB-C$^{5}$ & 78.5 & 80.4 & 78.4 & \textbf{82.4}\\
			\hline
			\hline
		\end{tabular}
	\end{center}

\end{table}

\begin{table}[!t]
	\begin{center}
	
	\caption{F1-score with setting 1 on Thai word
	segmentation task. 
% 	Note that the comparison is not direct because the models indicated with $^{*}$, used the dataset with different training and testing split.
	}
	\label{table2}
		\begin{tabular}{l|c}
			\hline
			\hline
			Models (\%)  & Novel\\
			\hline
			SGB-A$^{12}$ & 80.1\\
			SGB-C$^{12}$ & 79.2 \\
			SLM$^{12}$ & 79.6 \\
			PHYSMM & \textbf{82.1} \\
			nVBE & \textbf{82.1}\\
			\hline
			\hline
			
		\end{tabular}
	\end{center}

\end{table}

\begin{table}[!t]

	\begin{center}

	\caption{F1-score with setting 3 on Chinese word segmentation task}
		\label{table3}
		\begin{tabular}{l|c|c|c|c}
			\hline
			\hline
			Models (\%)  & CITYU & MSR & PKU & AS \\
			\hline
			nVBE & 77.5 & 78.2 & 77.9 & 72.9 \\
			ESA  & 74.9 & 78.4 & 77.4 & 77.9 \\
			Joint & -/- & 81.7 & 81.1 & -/-\\
			SGB-A$^{3}$  & 79.5 & 80.6 & 80.4 & 81.9\\
			SGB-C$^{3}$  & 77.6 & 81.1 & 80.4 & 79.7\\
			SGB-A$^{4}$  & 80.3 & 80.6 & 80.3 & 82.7\\
			SGB-C$^{4}$  & \textbf{80.5} & 81.7 & 81.0 & 82.3\\
			SGB-A$^{5}$  & 78.3 & 80.1 & 79.1 & 82.8\\
			SGB-C$^{5}$  & 80.2 & \textbf{82.3} & \textbf{81.2} & \textbf{83.5}\\
			\hline
			\hline
		\end{tabular}
	\end{center}

\end{table}

\begin{table}[!t]
	\begin{center}

		\caption{F1-score with setting 4 on Chinese word segmentation task}
			\label{table4}
		\begin{tabular}{l|c|c|c|c}
			\hline
			\hline
			Models (\%)  & CITYU & MSR & PKU & AS\\
			\hline
			nVBE & 76.7& 81.3  & 80.0 & 76.6 \\
			ESA  & 76.0 & 80.1 & 77.8 & 78.5 \\
			NPY$^{2}$  & 82.4 & 80.2 & -/- & -/- \\
			NPY$^{3}$   & 81.7 & 80.7 & -/- & -/-\\
			PHYSMM & \textbf{82.6} & 82.9 & \textbf{81.6} & -/- \\
			MCA  & 77.9 & 82.4 & 80.7 & 80.6\\
			
			SLM$^{2}$  & 78.2 & 78.5 & 80.2 & 79.4\\
			SLM$^{3}$  & 80.5 & 79.4 & 79.8 & 80.3\\
		    SLM$^{4}$ & 79.7 & 79.0 & 79.2 & 79.8\\
		    
			SGB-A$^{3}$  & 79.5 & 82.7& 80.9 & 81.7\\
			SGB-C$^{3}$  & 80.7 & 83.1& \textbf{81.6} & 82.0\\
			SGB-A$^{4}$  & 80.5& 81.7 & 79.4 & 83.0\\
			SGB-C$^{4}$  & 81.2& 83.6 & 81.5 & \textbf{83.9}\\
			SGB-A$^{5}$ & 78.7& 82.6  & 79.9 & 82.3\\
			SGB-C$^{5}$ & 79.8& \textbf{83.7}  & 80.8 & 83.8\\
			\hline
			\hline
		\end{tabular}
	\end{center}

\end{table}
\subsection{Disambiguation Analysis}
Segmentation ambiguity is a very important factor influencing accuracy of word segmentation systems \citep{huang2007chinese}.
We take Chinese word segmentation as an example, there are two types of word boundary ambiguities: combination ambiguity and overlap ambiguity.\\

\noindent
\textbf{Combination ambiguity } 
For any two  character sequences in a sentence, denoted by $x$ and $y$, if $xy$ can be combined together to form a word, yet at the same time $x$ and $y$ also can function as a word respectively, then word boundary ambiguity exists in these two character sequences.\\

\noindent
\textbf{Overlap ambiguity }
If there is a word boundary ambiguity between $xy$ and another character sequence that precedes or follows them, say $z$, and $xyz$ can be grouped into either $xy \, \, z$ or $x \, \, yz$, then there exists an overlap ambiguity.\\

\begin{figure}
	\centering
	\includegraphics[width=0.5\textwidth]{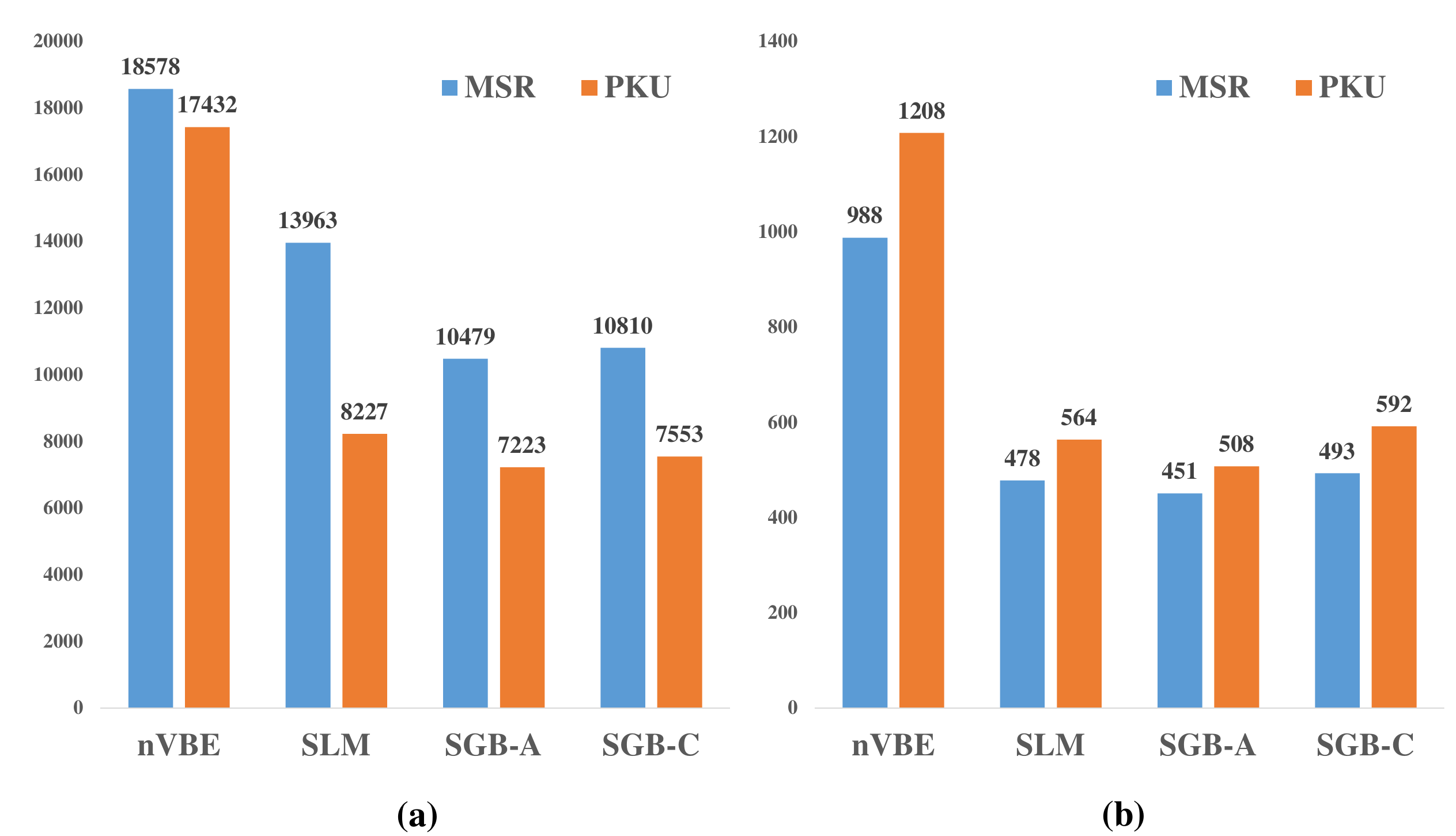}
	\caption{Segmentation error analysis. (a) The errors of different models caused by the combination ambiguity on MSR and PKU datasets. (b) The errors of different models caused by the overlap ambiguity.}
	\label{fig:ambiguity}
\end{figure}

We counted two kinds of ambiguity error made by our models on MSR and PKU datasets (the vocabulary is obtained from the gold test set) and compared SGB-A$^{3}$, SGB-C$^{3}$ with SLM$^{3}$ and nVBE. The result is shown in Figure \ref{fig:ambiguity}.

\section{Discussion}
\subsection{Comparison}
As shown in Table \ref{table1}, \ref{table2}, \ref{table3}, and \ref{table4}, our models outperform other approaches almost with every setting across different datasets for Chinese word segmentation.
Under the setting 4, we surpassed state-of-the-art by $3.3\%$ increase in F1-score on AS and $0.8\%$ increase on MSR, and achieved the compatible performance on PKU. We achieved new state-of-the-art under the setting 3 on CITYU and the decent results under the setting 4. 
Under the fully unsupervised situation (i.e. the setting 1), our models still perform well, and achieved more than 80\% accuracy on all the datasets.
Our model is evaluated on Thai that have longer word length in average than Chinese, and we also achieved the comparable result comparing to others.

Because the concept of words is not well defined in Chinese, different datasets were constructed following the different guidelines. \citet{huang2007chinese} estimated the the lowest consistency rate $0.848$ between four datasets which can be regarded as the upper bound for unsupervised Chinese word segmentation models. The result of our model is very close to this ceiling.

We found that our combining model achieve higher F1-scores than the average model on the most cases, while the averaging one show to be the better in word disambiguation.
Combining model leads to more fine-grained segmentation results which helps to isolate the function words from the other words. When the maximum length of words $T$ increases, the averaging model will suffer from the problem of function word adhesion (the function words can not be properly isolated). In contrast, the combining model always segments $xyz$ into $x \, \, y \, \, z$ rather than $xy \, \, z$ or $x \, \, yz$. Therefore, it makes the combining model hard to deal with the problem of word disambiguation because it tends to produce the results agreed by the both language models.

\subsection{Error Analysis}
%错误分析最好拿引言中的例子，来阐述为什么两种方法（combine 也要提到消灭歧义的面临的问题）能解决辅助词问题，average 获得了更精准的概率估计，以解决了其他方法不能解决的问题（这里一定要举例子，）。
% \begin{figure}
% 	\centering
% 	\includegraphics[width=0.5\textwidth]{error_2.pdf}
% 	\caption{Top-10 frequent characters occurred in wrong segmentation.}
% 	\label{Figure4}
% \end{figure}
The bidirectional nature of our models allows  them leverage the post-text information to determine the left boundary of the candidate segments. We take two sentences as examples: ``我从小学唱歌" (I started learning to sing when I was young), and ``我从小学毕业" (I graduated from primary school). The segment ``从小学" can be clearly split from  both sides, but there is an overlap ambiguity inside. Since segment "从小学" in these sentences have the same preceding text ("我"), no matter what the following word is (keep the boundary on right side), the one-side model SLM will split the segment in the same way which is determined by the result of comparing $p(\mbox{从小}|\mbox{我})p(\mbox{学}|\mbox{我从小})$ with $p(\mbox{从}|\mbox{我})p(\mbox{小学}|\mbox{我从})$. However our two-side model will refine the results by introducing the generation probability of segments given the context from the right side.  The occurrence of ``唱歌" (sing) will increase the probability of that ``学" (learn) is identified as a single word while the words ``毕业" (graduate) and ``小学" (primary school) are often occur together which establishes the left word boundary of ``小学". It is in accordance with the idea of branch entropy that uses the entropies from both sides to segment words. In addition, although our combining model can not handle the overlap ambiguity, the post contexts help to better segment the previous words. 

% We investigated the segmentation errors and count their character frequency occurred in all wrong segmentations, and divide errors into three parts: function words, numbers and others.
% Here we discuss the first two errors.

We find that the error of function word recognition is one of the main drawbacks of many existing approaches. Function words occur more frequently than the common words, and could be assigned with a higher probability by the language model, which causes these words easily to be attached with other words. Our combining model is able to handle such problem thank to the bi-directional language models. It can successfully establish the boundaries between the function words and next words.

Thai has less character types and longer word length in average than Chinese, which makes the models hard to accurately approximate the generation probabilities. As to our combining model, it tends to get fine-grated results that are not optimal for Thai. We leave it for future studies. Although our models did not define new state-of-the-art on Thai, we still achieved the comparable results on the Novel dataset.

\section{Conclusion}
% Chinese word segmentation remains an extremely difficult problem. The fact is that there is no existing approach that is able to reliably segment unfamiliar types of texts before fine-tuning with massive training data, not to mention the open texts mixed with the texts from different topics, genres, and regions. No matter how advanced the supervised segmentation has been, unsupervised approach still playing an important role in word segmentation since new word creation never stops and the lexicon are never sufficient.
We have described a novel unsupervised segmentation model augmented with bi-directional neural language models.  
Assuming that a good segmentation generally contributes to increase the generation possibility of sentences, the model's parameters are trained by maximizing such sentence generation probability conditioned on all the possible segmentation.
The bi-directional nature of networks allows us to capture the semantic-rich context features for each possible segment, aiming to better reconcile word boundary ambiguities. Two decoding algorithms were also proposed to combine the context features from both directions. One is to produce the segmentation by taking the average of segment's probabilities estimated by the forward and backward language models. Another is to generate the segmentation by merging their results produced separately by the two language models.The former revealed better in word boundary ambiguity reconciliation, while the latter achieved the higher overall performance. Experimental results  demonstrated that our context-sensitive unsupervised segmentation model achieved state-of-the-art on multiple Chinese data sets, and the comparable result on Thai in which the words are formed by more characters in average than those in Chinese.

% Based on Bayesian point of view, we proved that the sequence generation probability can be decomposed into the probability of its components (or segments) recursively, and thus a variant of dynamic programming algorithm can be used to obtain the optimal segmentation.

% Our framework is implemented and tested on the Chinese word segmentation (CWS) tasks, and the parameters are learned by maximizing the generation probability of sentences conditioned on all the possible segmentation of the sentences, assuming that good segmentation generally will contribute to increase the generation possibility of their corresponding sentences.

% Unsupervised approach provides us a possible ultimate solution. The experimental results on the second SIGHAN bakeoff show that our model achieves consistently higher performance over the five representative competitors, highlighting the potential of the proposed framework.

% It would be interesting to see how well our framework can be used into other tasks, such as  chunking and morphological segmentation. It is also possible to carefully optimize the neural network architecture and improve the training algorithm for those tasks.
\bibliography{emnlp-ijcnlp-2019}
\bibliographystyle{acl_natbib}

\end{CJK}
\end{document}